\documentclass{article}
\pdfoutput=1
\usepackage{ijcai19}

\usepackage{times}
\usepackage{soul}
\usepackage{url}
\usepackage{comment}
\usepackage{array}
\usepackage{multirow}
\usepackage{graphicx}
\usepackage{booktabs}
\usepackage{subfigure}
\usepackage{setspace}
\usepackage[hidelinks]{hyperref}
\usepackage[utf8]{inputenc}
\usepackage[small]{caption}
\usepackage{graphicx}
\usepackage{amsmath}
\usepackage{amssymb}
\usepackage{booktabs}
\usepackage{algorithm}
\usepackage{algorithmic}
\title{The Tradeoff Between Privacy and Accuracy in Anomaly Detection \\Using Federated XGBoost}

\author{
Mengwei Yang$^1$
\and
Linqi Song$^*{}^1$\and
Jie Xu$^2$\And
Congduan Li$^3$\and
Guozhen Tan$^4$
\affiliations
$^1$City University of Hong Kong\\
$^2$University of Miami\\
$^3$Sun Yat-sen University\\
$^4$Dalian University of Technology
\emails
mengweiy.ray@gmail.com,
linqi.song@cityu.edu.hk,
jiexu@miami.edu,
licongd@mail.sysu.edu.cn,
gztan@dlut.edu.cn
\thanks{The corresponding author is L. Song. This  work was  supported  in  part  by  the City University of Hong Kong Grant (No. 7200594), the Hundred Talents Program of Sun Yat-sen University (Project No. 76150-18841214), and the Key  Program  of  the National Natural Science Foundation of China under Grant No. U1808206.}
}

\begin{document}

\maketitle

\begin{abstract}
Privacy has raised considerable concerns recently, especially with the advent of information explosion and numerous data mining techniques to explore the information inside large volumes of data. In this context, a new distributed learning paradigm termed federated learning becomes prominent recently to tackle the privacy issues in distributed learning, where only learning models will be transmitted from the distributed nodes to servers without revealing users' own data and hence protecting the privacy of users.

In this paper, we propose a horizontal federated XGBoost algorithm to solve the federated anomaly detection problem, where the anomaly detection aims to identify abnormalities from extremely unbalanced datasets and can be considered as a special classification problem. Our proposed federated XGBoost algorithm incorporates data aggregation and sparse federated update processes to balance the tradeoff between privacy and learning performance. In particular, we introduce the virtual data sample by aggregating a group of users' data together at a single distributed node. We compute parameters based on these virtual data samples in the local nodes and aggregate the learning model in the central server. In the learning model upgrading process, we focus more on the wrongly classified data before in the virtual sample and hence to generate sparse learning model parameters. By carefully controlling the size of these groups of samples, we can achieve a tradeoff between privacy and learning performance. Our experimental results show the effectiveness of our proposed scheme by comparing with existing state-of-the-arts. 

\end{abstract}

\section{Introduction}

Nowadays, many giant internet companies, like Google, Amazon, and Alibaba, have established large scale information technology infrastructures to cope with the current huge data stream and to provide numerous services to customers. However, the large volume of data will also bring a number of serious privacy issues \cite{chen2012data} and computing problems. For example, in social networks like Facebook, there is a growing concern of privacy risk in collecting a large amount of users' private data, including various personal information, texts, pictures, and video data. Leveraging these large volumes of human data, these companies will utilize them to train various machine learning models for various data intensive applications. However, users can do little to protect their data. As such, in May 2018, the European Union has began to implement the General Data Protection Regulation (GDPR) to protect individual privacy \cite{voigt2017eu}, which is deemed the most important change in data privacy regulation in 20 years.

Even though in some areas, data can be shared between different companies or concentrated on some cloud servers, it still carries dramatic risks and transmission issues. On one hand, the transfer of private data between different parties makes it more likely to leak or to be hacked. On the other hand, the transmission of large amounts of data leads to inefficiency. In this context, the federated learning framework has been proposed and plays an indispensable role in solving these problems \cite{hard2018federated}. Instead of transmitting raw data, federated learning transmits pre-trained learning models from users to servers, while keeping the users data locally. Thus, the user privacy can be protected; computing resources in the user side can be efficiently utilized; and the communication cost is reduced.


Recently, federated learning has attracted broader attention and three categories was put forward in \cite{yang2019federated}, including horizontal federated learning, vertical federated learning and federated transfer learning. In \cite{cheng2019secureboost}, the {\em SecureBoost} was presented, which achieves vertical federated learning with a tree-boosting algorithm. 

In this work, we propose a horizontal federated XGBoost algorithm\footnote{\noindent https://github.com/Raymw/Federated-XGBoost} in anomaly detection with an application in detecting the fraudulence in bank credit card transactions; and study the trade-off behavior between the privacy preserving and the anomaly detection performance. Compared with SecureBoost, which was deployed in the vertical federated learning framework, our federated XGBoost is a horizontal federated learning algorithm where different data samples with all features are distributed among the distributed nodes. Though tree-boosting algorithm is also utilized in this work, this horizontal federated XGBoost is deployed in a totally different way. First, the biggest difference is the transfer of parameters. It is far from enough to only pass parameters $g_i$ and $h_i$ in horizontal federated XGBoost. Because when using tree-boosting algorithms, different nodes have to obtain instances of every feature so that the gain of split can be calculated and the split point can be acquired. Another key difference is to use a two step (data aggregation and federated update) method to preserve individual data's anonymity which we will describe later. Furthermore, the sparse federated update by focusing on wrongly classified data is utilized in federated XGBoost to improve the process of federated update. 

\begin{figure}[!htb]
\centering
\includegraphics[scale=0.45]{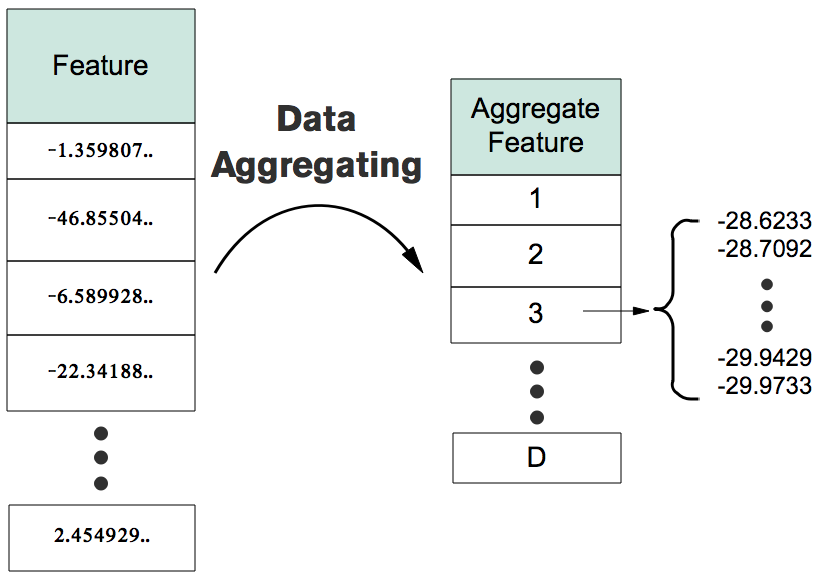}
\caption{The process of data aggregating in federated XGBoost framework.}
\label{data_aggregatingfig}
\vspace{-0.1in}
\end{figure}

In particular, to transfer features of users in a privacy and efficient way, our proposed two-step method is described as follows. The first step is {\bf Data Aggregation}: First of all, the privacy of users should be protected and thus users' information can't be passed directly. Hence, for the purpose of calculating the gain of split mentioned above, features of users are entailed. In this paper, for the consideration of protecting users' information, instead of directly transmitting all exact data in each feature, the original data in each feature is projected in an anonymous way by using modified K-Anonymity, shown in Figure \ref{data_aggregatingfig}, where a group of data samples have been mapped to a virtual data sample. The projection is implemented under every feature. Because while finding the split point, the purpose of original data for tree-boosting algorithm is to get the sequence under every feature. So, after passing the number of virtual data samples in each feature, the gain of split can be calculated.
Consequently, by doing this, not only the privacy of users will be protected, but also the tree-boosting model can decide the split point and build the tree.

A second step is the {\bf Federated Update}: In reality, the amount of data is quite large, it is inefficient to transfer all data and also not all data is valuable for update. In that case, it is necessary to filter data so as to better update models. Though the tree-boost model can implement well in prediction by building trees, there are still many instances that cannot be classified correctly. Hence, in this paper, wrongly classified instances will be processed with more focus and then be transferred to server for federated update. The reason is that firstly, these instances are more valuable and will help the model improve itself better. Also because the data used in anomaly detection is extremely unbalanced, the boosting algorithm can solve skewed problem in some degree and elevate the generalization ability of the model. Secondly, if the correctly classified data is not filtered, these instances will affect the process of split and the construction of trees in the process of federated update, which has an adverse impact on the improvement of the model.

We show a trade-off behavior between the detection accuracy and the the privacy measured in terms of k-anonymity. Through simulation experiments, we find a reasonable size of the number of virtual samples in data aggregation so that the privacy of users will be better protected and the learning performance in federated XGBoost will be reduced to a minimum. We show that our proposed algorithm achieves up to 5\% performance gains in terms of F1-core compared with existing state-of-the-art algorithms, while effectively keeping user privacy.


\section{Related Work}

\paragraph{Privacy-preserving and Federated Learning:}

The transfer of data will bring the problem of data leakage \cite{shokri2015privacy}. Consequently, decentralized methods (i.e., data is only stored locally) are used to process the data and then the risk of data leakage is reduced \cite{wang2010privacy}. Some works use encryption-based federated learning frameworks, like homomorphic encryption \cite{gilad2016cryptonets}. Homomorphic encryption means certain operations can be acted directly on encrypted data without decrypting it. However, homomorphic encryption has its disadvantages. Take Paillier-based encryption schemes as an example, the cost of generating threshold decryption keys is very high \cite{bonawitz2017practical}.

Federated learning is a new distributed learning paradigm proposed recently to utilize the user-end computing resources and preserve user's privacy by transmitting only model parameters, instead of raw data, to the server \cite{1812.03337,1901.08277}. In federated learning, a general model will be firstly trained, and then the model will be distributed to each node acting as a local model \cite{yang2019federated,mcmahan2016communication,konevcny2016federated}. Three  categories  was  put  forward  in \cite{yang2019federated}, including horizontal federated learning, vertical federated learning and federated transfer learning. The federated secure XGBoost framework using vertical federated learning was proposed in \cite{cheng2019secureboost}.

In contrast, in this work, we firstly preprocess data where instances can be merged together to learn an aggregate gradient such that the communication and computation cost will be significantly reduced. Next, we generate local models and aggregate those models in the central server to update the original model. By doing so, we can show a tradeoff between the privacy of the user and the learning performance.

\paragraph{Anomaly Detection:}

Anomaly detection \cite{patcha2007overview} is the identification of events or observations that do not match the expected pattern or other items in the dataset (i.e., outliers) during data mining. Outliers can be divided into point exceptions, context exceptions, and collective exceptions \cite{agrawal2015survey}. Anomaly detection methods include SMOTE algorithm \cite{chawla2002smote} and various machine learning models, such as K-Nearest Neighbors algorithm \cite{liao2002use}, Random Forest \cite{zhang2008random}, Support Vector Machine (SVM) \cite{li2003improving}, Gradient Boosting Classification Tree (GBT) \cite{krauss2017deep}, XGBoost \cite{chen2016xgboost}, and deep learning neural network models \cite{mukkamala2002intrusion}. In this paper, a point exception of fraud detection in credit card transactions will be focused and the dataset of credit card transactions will be used to train the model.

\section{Problem Formulation}
\begin{figure}[!htbp]
\centering
\includegraphics[scale=0.4]{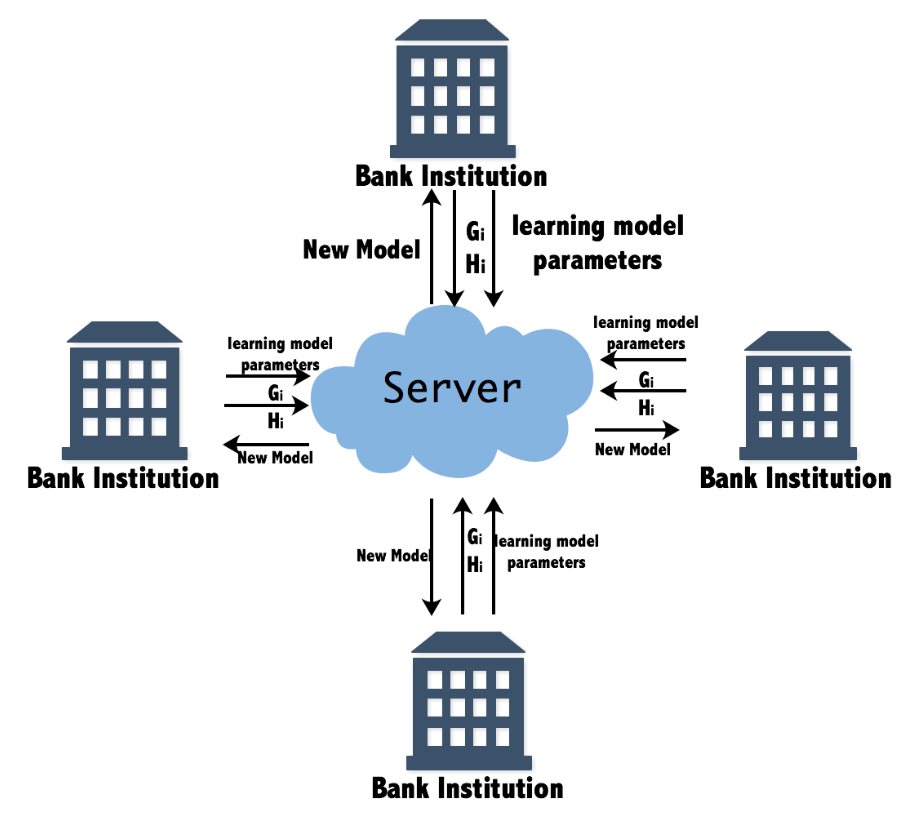}
\caption{The illustration of federated learning framework.}
\label{fig.transmit}
\vspace{-0.05in}
\end{figure}

We consider the federated learning  in an anomaly detection problem as follows. There are $D$ distributed nodes, e.g., bank institutions, denoted by $1,2,\ldots,D$. In each node $j$, the local data $X_j$ is given with $n_j$ data instances (i.e., data examples) and $d$ features, i.e., $ X_j =\left \{ (x_{i},y_{i})\right \} (|X_j| = n_j, x_{i} \in \mathbb{R}^{d},y_{i} \in \mathbb{R})$. We denote the union of all local data to be $X$. There is a center server node to aggregate the learning model. The entire system architecture is shown in Fig.~\ref{fig.transmit}. 

The federated learning process is as follows. First, the local nodes preprocess the data (the local data) and send some learning model parameters to the server. Second, the server will integrate those received model parameters and obtain a new global model. The new model will be transmitted to local nodes as well. This process will be iterated over time again and again to train a sufficiently good model. Through this learning process, the local data do not need to be exchanged and the user privacy is protected.  

In the anomaly detection problem, the data is often very unbalanced; namely, in most of the cases, the data points are normal (i.e., most samples with a label $0$), and in rare cases, the data points are abnormal (i.e., rear samples with a label $1$). In this case, the extremely skewed data on each node can not represent for the overall distribution. In that case, each node needs to share data in the whole federated learning framework, which can help to improve the existed model in each node. The goal is to train a federated learning model to detect the abnormalities using the federated learning system as described above. 

In this paper, we ask the question: {\bf if the local nodes choose different learning model parameters to be transmitted to the server, then what is the tradeoff between the machine learning performance and the user privacy.} 

\paragraph{Performance Metrics}
Next, we describe the performance metrics of the anomaly detection and the user privacy.

\paragraph{$\bullet$ Modified $k$-Anonymity} $k$-anonymity is a property possessed by certain anonymized data, where one cannot distinguish a user out of $k$ candidates from this set of data \cite{sweeney2002k}. Here, since local nodes transmit learning model parameters to the server, we define a \emph{modified $k$-anonymity} metric as the privacy property that we cannot distinguish a user out of other candidates from the transmitted learning parameters instead of a set of data.

\paragraph{$\bullet$ Measurement for Unbalanced Data}
In anomaly detection, the data is unbalanced and usually it is not a good idea to simply use accuracy to measure the learning performance as classifying all data into the normal category will result in a sufficient high accuracy. For example, in the bank credit card fraudulence detection dataset that we use in experiments, fraud cases account for only $0.172\%$ of the total data. In actual banking transactions, fraudulent transactions are still the minority \cite{phua2004minority}. Consequently, we will use the confusion matrix, F1-score and AUPRC (the area under Precision-Recall curve) to measure the anomaly detection performance. An illustration of these concepts are shown in Fig.~\ref{fig.confusionMatrix}. 

\begin{figure}
\centering
\newcommand\MyBox[2]{
  \fbox{\lower0.75cm
    \vbox to 1.2cm{\vfil
      \hbox to 1.2cm{\hfil\parbox{1.4cm}{#1\\#2}\hfil}
      \vfil}%
  }%
}
\noindent
\renewcommand\arraystretch{1.1}
\setlength\tabcolsep{0pt}
\begin{tabular}
{c >{\bfseries}r @{\hspace{0.7em}}c @{\hspace{0.4em}}c @{\hspace{0.7em}}l}
  \multirow{10}{*}{\rotatebox{90}{\parbox{1.1cm}{\bfseries\centering Actual\\ value}}} & 
    & \multicolumn{2}{c}{\bfseries Prediction outcome} & \\
  & & \bfseries p & \bfseries n & \bfseries total \\
  & p$'$ & \MyBox{True}{Positive} & \MyBox{False}{Negative} & P$'$ \\[2.4em]
  & n$'$ & \MyBox{False}{Positive} & \MyBox{True}{Negative} & N$'$ \\
  & total & P & N &
\end{tabular}

\caption{Illustration of confusion matrix. $TP$ = predicted positive and it is true; $FP$ = predicted positive but it is negative; $FN$ = predicted negative but it is positive; $TN$ = predicted negative and it is true. For the Precision-Recall curve, the X-axis is Recall and Y-axis is Precision. \protect\begin{math}Precision = TP / (TP + FP); Recall = TP / (TP + FN) \protect\end{math}; The AUPRC is mainly used for the judgment of unbalanced dataset. The F1-score is introduced to judge the result of the model prediction: F1-score \protect\begin{math} = 2*TP / (2*TP+FP+FN) \protect\end{math}.}
\label{fig.confusionMatrix}
\vspace{-0.15in}

\end{figure}


\section{The Federated XGBoost Framework}
In this section, we will talk about the federated XGBoost algorithm that we will use in the anomaly detection problem. We will first give a recap of the XGBoost algorithm, a general description of the federated XGBoost algorithm, and some specific design tailored to the anomaly detection problem.

\subsection{Preliminaries of XGBoost}
We give a brief overview of the XGBoost algorithm, and one can refer to \cite{chen2016xgboost} for more details.

For machine learning problems, such as the classification or regression, given a dataset $\{(x_i,y_i)\}$ of $n$ examples and $d$ features, the goal is to train a learning model with parameters $\theta$ to minimize the objective loss function as follows 

\begin{equation} 
Obj(\theta )=L(\theta )+\Omega(\theta )
\end{equation}
where \begin{math} L(\theta ) \end{math} is the training loss and \begin{math}  \Omega(\theta ) \end{math} is the regularization term. For XGBoost algorithm, it utilizes $K$ regression/classification trees to predict the output, where the predicted output for the $i$-th data example is  

\begin{equation}
\hat { y }_i = \phi \left( x _ { i } \right) = \sum _ { k = 1 } ^ { K } f _ { k } \left( x _ { i } \right).
\end{equation}

So for the objective function of XGBoost:
\begin{equation}L ( \phi ) = \sum _ { i } l \left( y _ { i } , \hat { y } _ { i } \right) + \sum _ { k } \Omega \left( f _ { k } \right)  \end{equation} 
where \begin{math}\Omega\left( f _ { k } \right)=\gamma T + \frac { 1 } { 2 } \lambda | | w | | ^ { 2 }\end{math} with component $w_j$ of $w$ being the score/weight on $j$-th leaf of the tree. Since the newly generated tree needs to fit the last predicted residual. So 
 \begin{math}\hat{y}_i\end{math} can be written as \begin{math}\hat{y}_i^{(t)}=\hat{y}_i^{(t-1)}+f_t(x) \end{math} for the $t$-th iteration. Also, take Taylor expansion of the objective as follows:

\begin{equation} { L } ^ { ( t ) } \simeq \sum _ { i = 1 } ^ { n } \left[ l \left( y _ { i } , \hat { y }_i ^ { ( t - 1 ) } \right) + g _ { i } f _ { t } \left( \mathbf { x } _ { i } \right) + \frac { 1 } { 2 } h _ { i } f _ { t } ^ { 2 } \left( \mathbf { x } _ { i } \right) \right] + \Omega \left( f _ { t } \right) \end{equation}
where \begin{math}g_i=\partial_{\hat{y}_i^{(t-1)}}\;l(y_i, \hat{y}_i^{(t-1)}),  h_i=\partial_{\hat{y}_i^{(t-1)}}^2 \; l(y_i, \hat{y}_i^{(t-1)})\end{math}. Here,  \begin{math}g_i\end{math} and \begin{math}h_i\end{math} represent the first and second gradient statistics of the loss function. By using greedy algorithm to search the best split which aims to maximize the learning gain at each iteration:

\begin{equation} 
Gain=\frac{1}{2}\left[\frac{G_L^2}{H_L+\lambda}+\frac{G_R^2}{H_R+\lambda}-\frac{(G_L+G_R)^2}{(H_L+H_R)+\lambda}\right]-\gamma 
\end{equation}

\begin{equation}w_j^* = -\frac{G_j}{H_j+\lambda} \end{equation}

where \begin{math} G_L=\sum_{i\in I_L}g_i\end{math}, \begin{math}G_R=\sum_{i\in I_R}g_i\end{math}, \begin{math}H_L=\sum_{i\in I_L}h_i\end{math}, \begin{math}H_R=\sum_{i\in I_R}h_i\end{math}. Here, $I_L$ and $I_R$ represent the left and right sets of data sample indices. The equation of $Gain$ is used for evaluating the split point and \begin{math} w_j^* \end{math} denotes the weight of leaf. When searching for the best split point, instances'  \begin{math}g_i\end{math} and \begin{math}h_i\end{math}  in the left and right space will be calculated for getting the value of $Gain$. 

Without loss of generality, we consider a particular logistic loss function $l(y_i, \hat{y}_i^{(t-1)})=y_i\ln(1+e^{-\hat{y}_i})+(1-y_i)\ln(1+e^{\hat{y}_i})  $. So that 
$$g_i=\frac{1}{1+e^{-\hat{y}_i^{(t-1)}}}-y_i, h_i=\frac{1}{1+e^{-\hat{y}_i^{(t-1)}}}*(1-\frac{1}{1+e^{-\hat{y}_i^{(t-1)}}}).  $$ 
These parameters will be used for parameters passing in the federated learning framework. 

\subsection{Federated XGBoost}
In the federated learning framework, to implement the XGBoost algorithm, one simple idea is to calculate the parameters $g_i$ and $h_i$ of each data sample at each local node, and then transmit these parameters to the center server to determine an optimal split. 

Note that in vertical data partition \cite{cheng2019secureboost}, different node holds one part of the same instance, so that by only passing parameters between each other, the model can make predictions in cooperation with other nodes. In this paper, we consider the \emph{horizontally partitioned data} in different local nodes, which means that data provided in different node have the same feature dimension and one node holds all features of an instance. It is not easy to update other models in different nodes if only transmitting parameters ($g_i$ and $h_i$). Though by averaging the parameters of different models does help, it still cannot ameliorate the model a lot in each node.

Here, instead of simply transmitting model parameters $g_i$ and $h_i$, we make two revisions that are tailored to the specific anomaly detection setting: a \emph{data aggregation} process and a \emph{sparse federated update} process. 

First, in the data aggregation process, we map a range of data samples that are close to each other into a virtual data sample (or a cluster of samples). Taking into account each virtual data sample as a new data sample $I_1$, we sum up the $g_i$s and $h_i$s ($i\in I_1$) in this cluster to obtain the $g_{I_1}$ and $h_{I_1}$ for this virtual data sample. We then transmit parameters corresponding to these virtual samples to the central server to train the model. However, when new learning models are obtained, these data samples will calculate their losses and parameters $g_i$ and $h_i$ separately. By controlling the size of the virtual sample, we can achieve a tradeoff between learning performance and the privacy in terms of modified $k$-anonymity.

In our modified $k$-anonymity, instances in every feature will be mapped into different virtual data nodes. As shown in Figure \ref{data_aggregatingfig}, we use the sequence number of virtual data nodes (from $1,2,\ldots,D$) to represent a cluster of samples’ exact information, which means individual values in every feature are replaced by a new category. Since every virtual data node represents a range of samples’ values, samples inside every node are anonymous and attackers cannot distinguish a user out of other candidates. Also, even though attackers get instance $A$’s exact values, they still cannot acquire instance $A$’s other sensitive information because of the adoption of a new category. Therefore, our modified $k$-anonymity in this work can protect users’ privacy in an anonymous way.

Second, to further improve the communication efficiency and the anomaly detection performance, in the sparse federated update process, we will focus on tackling these wrongly classified instances. Our assumption is that at iteration $t$, for most data samples, the function $\sum _ { k = 1 } ^ { t } f _ { k } \left( x _ { i } \right)$ will give sufficient accurate estimations. Therefore, at iteration $t+1$, we will just focus on the \emph{wrongly} classified samples. Therefore, we will calculate the $g_{I_1}$ and $h_{I_1}$ for cluster $I_1$ by summing up only $g_i$s and $h_i$s of those samples $i$ that are wrongly classified from the learning model thus far. 

The details of our proposed federated XGBoost algorithm are shown in Algorithms~\ref{alg:1} and \ref{alg:2}, where the first one is for local node to compute local models and the second one is for the server to aggregate the global model. 

\begin{algorithm}[!t]
	\renewcommand{\algorithmicrequire}{\textbf{Input:}}
	\renewcommand{\algorithmicensure}{\textbf{Output:}}
	\caption{The Federated XGBoost Framework (local node)}
	\label{alg:1}
	\begin{algorithmic}[1] 
		\REQUIRE $I$, Instance space of current node
		\REQUIRE $d$, Feature dimension
		\REQUIRE $v$, Mapping dimension of sequence in every feature
		\REQUIRE $\{g_{i},h_{i}\}_{i\in l} $,  $\{g_{i}^{L},h_{i}^{L}\}_{i\in l} $
			\FOR{$n=0$ to $d$}
			\STATE $//$ Original instances of model
			\STATE Propose $S_{n}$ =\{$s_{n1}$, $s_{n2}$,... $s_{nv}$\} by sequence on feature $n$.
			\STATE $G_{nv}$ =$\sum_{i\in \{ i|s_{n,v} \geq x_{i,n}>s_{n,v-1} \}} $ $g_{i}$
			\STATE $H_{nv}$ =$\sum_{i\in \{ i|s_{n,v} \geq x_{i,n}>s_{n,v-1} \}} $ $h_{i}$
			\ENDFOR
			\FOR{$n=0$ to $d$}
			\STATE $//$ New wrongly classified instances
			\STATE Propose $S_{n}^{L}$ =\{$s_{n1}^{L}$, $s_{n2}^{L}$,... $s_{nv}^{L}$\} by sequence on feature $n$.
			\STATE $G_{nv}^{L} $ =$\sum_{i\in\{i|s_{n,v}^{L} \geq x_{i,n}^{L}>s_{n,v-1}^{L}\}}$ $g_{i}^{L}$
			\STATE $H_{nv}^{L} $ =$\sum_{i\in\{i|s_{n,v}^{L} \geq x_{i,n}^{L}>s_{n,v-1}^{L}\}}$ $h_{i}^{L}$
			\ENDFOR	
			\STATE $//$ Integrate instances 
			\STATE $G_{nv}^{N}$ $\leftarrow$ $G_{nv}^{L}$ + $G_{nv}$ 
			\STATE $H_{nv}^{N}$ $\leftarrow$ $H_{nv}^{L}$ + $H_{nv}$

		\ENSURE  $G_{nv}^{N}$, $H_{nv}^{N}$
	\end{algorithmic}  
\end{algorithm}

\begin{algorithm}[!t]
	\renewcommand{\algorithmicrequire}{\textbf{Input:}}
	\renewcommand{\algorithmicensure}{\textbf{Output:}}
	\caption{The Federated XGBoost Framework (Server)}
	\label{alg:2}
	\begin{algorithmic}[1] 
		\REQUIRE $I$, Instance space of current node
		\REQUIRE $d$, Feature dimension
		\REQUIRE $v$, Mapping dimension of sequence in every feature
		\REQUIRE $G_{nv}^{N}$, $H_{nv}^{N}$ from local node
			
			\FOR{$n=0$ to $d$} 
			\STATE $//$ Enumerate all features
			\STATE $G$ $\leftarrow$ $\sum_{k=0}^v$ $G_{nk}^{N}$, $H$ $\leftarrow$ $\sum_{k=0}^v$ $H_{nk}^{N}$
			\STATE $G_{L}$ $\leftarrow$ $0$,\ $H_{L}$ $\leftarrow$ $0$
				\FOR{$k=0$ to $v$} 
			    \STATE $//$ Enumerate the value of Gain in each split point
				\STATE $G_{L}$ $\leftarrow$ $G_{L}$ + $G_{nk}^{N}$ ,\ $H_{L}$ $\leftarrow$ $H_{L}$ + $H_{nk}^{N}$
			    \STATE $G_{R}$ $\leftarrow$ $G$\ -\ $G_{L}$ ,\ \ $H_{R}$ $\leftarrow$ $H$\ -\ $H_{L}$
				\STATE score $\leftarrow$ $max(score, \frac{G_L^2}{H_L+\lambda}+\frac{G_R^2}{H_R+\lambda}-\frac{G^2}{H+\lambda})$
				\ENDFOR
			\ENDFOR
		\ENSURE  The direction of split in federated learning settings
	\end{algorithmic}  
\end{algorithm}

\section{Experimental Results}
In this section, we present our experimental results over a real dataset for credit card fraud detection. We will first describe the characteristics of the dataset and then show our algorithm's performance (we use F-XGBoost to represent our algorithm) compared with other existing state-of-the-arts.

\textbf{Credit Card Fraud Dataset\footnote{\noindent https://www.kaggle.com/mlg-ulb/creditcardfraud}: }This dataset contains transactions generated by credit card and it has 492 frauds and 284807 transactions, which is greatly unbalanced. It is a dataset that contains 30 features. Features V1, V2, ... V28 are the principal components obtained with PCA and only two features, Time and Amount, are kept original.

\textbf{Experimental Setting:}
We split the dataset into two parts: one for basic model training and the other for simulating the situation of newly acquired and also wrongly classified instances. Nearly 1/5 of the dataset (59875 tuples) is used for updating existed models in the federated learning settings and 4/5 of the dataset will be divided into testing (45569 tuples) and training data (179363 tuples). In the experiment, the XGBoost\footnote{\noindent https://github.com/dmlc/xgboost}, GBDT\footnote{\noindent http://scikit-learn.org/stable/modules/generated/sklearn.ensemble. GradientBoostingClassifier.html} and Random Forest\footnote{\noindent
http://scikit-learn.org/stable/modules/generated/sklearn.ensemble.
Random ForestClassifier.html} are utilized for performance comparisons and the parameter settings of XGBoost and federated XGBoost framework are the same. Learning rate is set as 0.1, maximum depth is 4. 

\begin{figure}[!htb]
\centering
\includegraphics[scale=0.37]{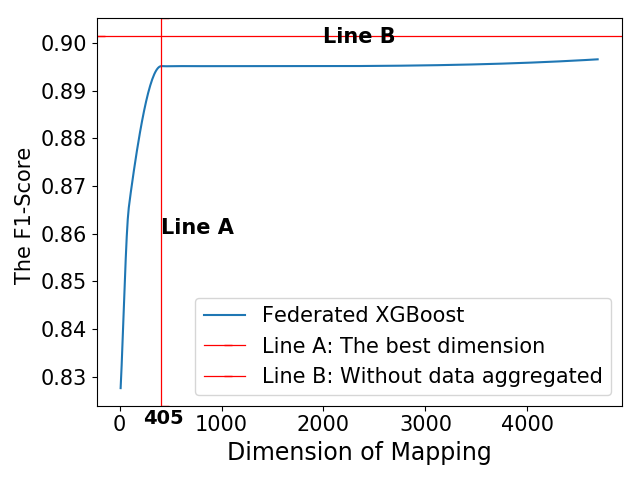}
\caption{The F1-Score of Federated XGBoost Framework in different dimensions of mapping.}
\label{fig.mapping}
\end{figure}

\textbf{Experimental Results:}
When the virtual cluster's size is larger, the protection of privacy will be better. However, it is a trade-off between the cluster size and the accuracy. In the original dataset, the number of samples (sample clusters) in each feature is 275665. As it shown in Figure \ref{fig.mapping}, Line B means in the original 275665 dimension, the F1-Score of federated XGBoost framework will be 0.901408. In Figure \ref{fig.mapping}, we can see a tradeoff behavior between the learning performance and the privacy, where the horizontal axis represents the number of clusters and the vertical axis represents the F1-score. We show that with the increase in sample clusters, the privacy preserving ability is decreased while the learning performance is improved.   
In Figure \ref{fig.mapping}, we can see that when the number of clusters is 405, the F1-Score is 0.895105.

In Table 1, the high accuracy of all models can be noticed, which means the evaluation parameters such as accuracy is not appropriate to fully evaluate models for this extremely unbalanced dataset. With emphasis, some good evaluating parameters like F1-Score, AUC and AUPRC can be deployed. 

\begin{figure}[!htb]
\centering
\includegraphics[scale=0.37]{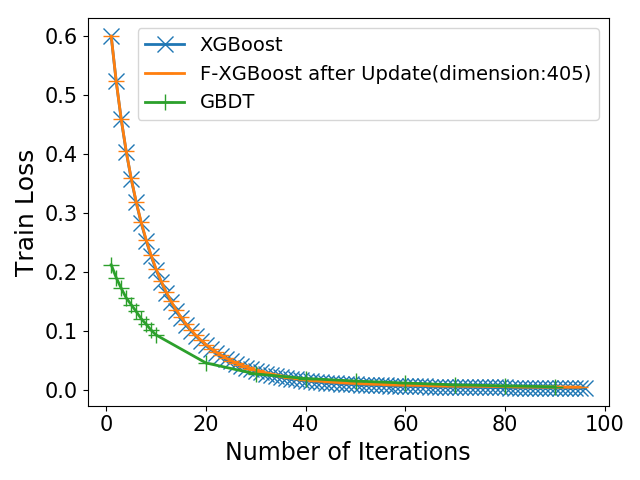}
\caption{Train Loss of Federated XGBoost Framework and XGBoost}
\label{fig.loss}
\end{figure}

\begin{figure}[!htb]
\centering
\includegraphics[scale=0.49]{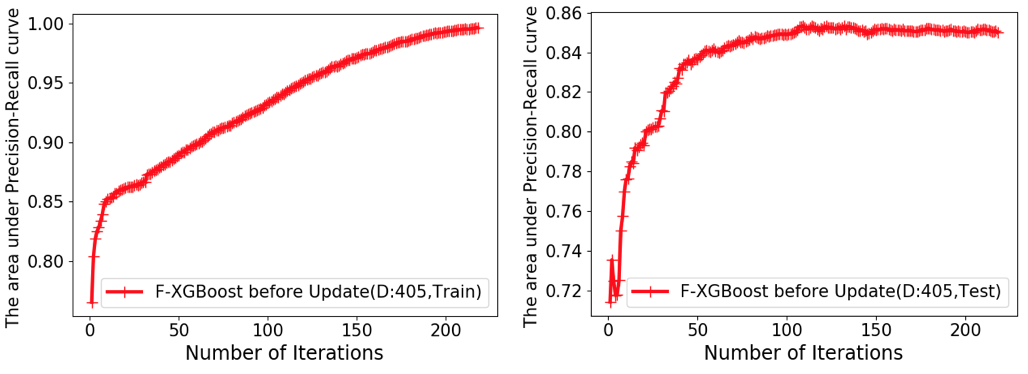}
\caption{AUPRC (The Area under Precision-Recall Curve) of The federated XGBoost framework before update; Left adopts Train set; Right adopts Test set.}
\label{fig.AUPRC1}
\end{figure}

\begin{figure}[!htb]
\centering
\includegraphics[scale=0.39]{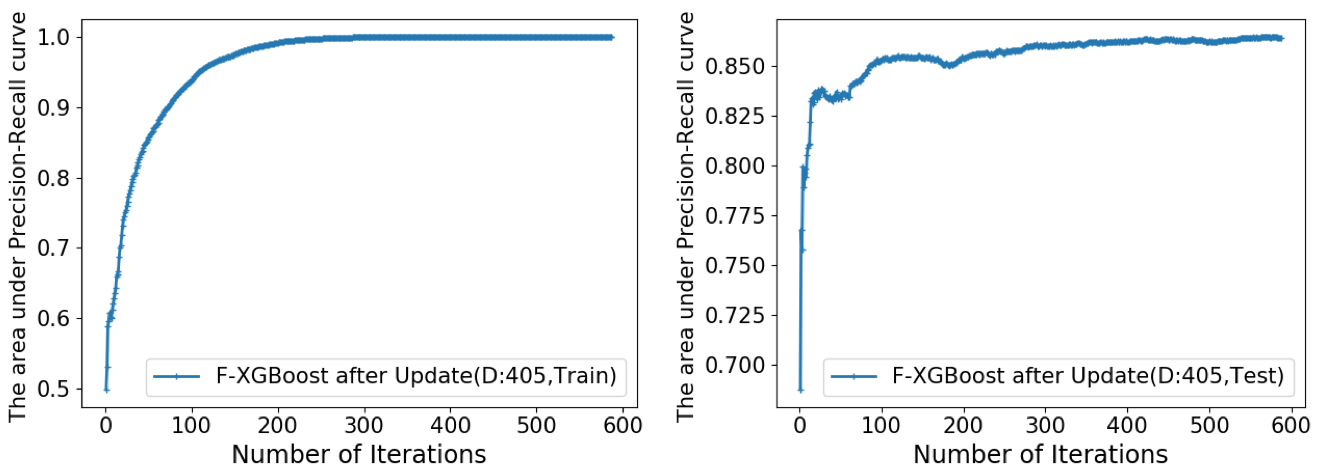}
\caption{AUPRC (The Area under Precision-Recall Curve) of The federated XGBoost framework after update; Left adopts Train set; Right adopts Test set.}
\label{fig.AUPRC2}
\end{figure}

From Table 1, we also show the AUC and F-1 curve for different algorithms. Compared with Random Forest, GBDT and the federated XGBoost before Update (Original Dimension), the updated federated XGBoost framework (Original Dimension) has obviously good performance over F1-Score. We can see that the proposed algorithm outperforms existing methods by up to 3.4\% in AUC and 5\% in F1-score. For federated XGBoost, the dimension of 405 achieves a reasonable trade-off between privacy and accuracy, though the F1-score of federated XGBoost framework (Dimension:405) is 0.63\% lower than federated XGBoost framework (Original Dimension). Also, the AUPRC displays the improvement of federated learning model compared with itself. The AUPRC performance is shown in Figures \ref{fig.AUPRC1} and \ref{fig.AUPRC2} for training and test data sets. For the train loss in Figure \ref{fig.loss}, the learning curve of federated XGBoost framework shows how the learning works and we can see that the training loss of our proposed federated XGBoost decreases faster than the GBDT algorithm. 






%




\begin{table}[!htb]
\caption{Performance of the Federated XGBoost (F-XGBoost) Framework} 
\label{Table.1} 
\begin{center}
\begin{tabular*} {8.7 cm} {@{\extracolsep{\fill} }ccccccccc} 
\toprule
Model & Accuracy \\
\midrule	
GBDT                        &  0.9995\\
Ramdom Forest                        &  0.9995\\
F-XGBoost before update (Original Dimension)  &  0.9994\\
F-XGBoost before update (Dimension:405)  &  0.9996\\
F-XGBoost after update (Original Dimension)  &  0.9997\\
F-XGBoost after update (Dimension:405)  &  0.9997\\
\bottomrule
\toprule
Model & AUC \\
\midrule	
GBDT                        &  0.9700\\
Ramdom Forest                        &  0.9456\\
F-XGBoost before update (Original Dimension) &  0.9214\\
F-XGBoost before update (Dimension:405)  &  0.9641\\
F-XGBoost after  update (Original Dimension)   &  0.9748\\
F-XGBoost after update (Dimension:405)  &  0.9733\\
\bottomrule
\toprule
Model & F1-Score &  \\
\midrule	
GBDT                        &  0.8571\\
Ramdom Forest                        &  0.8591\\
F-XGBoost before update (Original Dimension)  &  0.8169\\
F-XGBoost before update (Dimension:405)  &  0.8652\\
F-XGBoost after  update (Original Dimension)  &  0.9014\\
F-XGBoost after update (Dimension:405)  &  0.8951\\
\bottomrule
\end{tabular*} 
\end{center}
\end{table} 

\section{Conclusion and Future Work}

In this paper, we proposed a federated XGBoost algorithm to solve the anomaly detection problem. We show a tradeoff behavior between the learning performance and the privacy. In experimental results, we show some reasonable working points which achieve a balance between the privacy and accuracy. Also, by comparing with other algorithms, the effectiveness of this federated XGBoost framework can be clearly found with up to 5\% performance gains. In our proposed federated XGBoost framework, we use two techniques, data aggregation and sparse federated update, to reduce the communication and computing cost while improving the anomaly detection ability. More importantly, the privacy of users is protected and through the process of data aggregation, the risk of users' information leakage is avoided.

In our future work, we will attempt to experiment and deploy differential privacy in federated XGBoost so as to better protect users' privacy. Also, it is believed that there are still a lot of details should be considered and more research should be done on federated learning to make it more significant.


\newpage
\bibliographystyle{named}

\end{document}